\documentclass[preprint,1p,12pt,authoryear]{elsarticle}

\biboptions{longnamesfirst}
\usepackage{booktabs}
\usepackage{amsmath}
\usepackage{amssymb}
\usepackage{multirow}
\usepackage{xcolor}
\usepackage{graphicx}
\usepackage{adjustbox}
\usepackage{float}

\newcommand{\RR}{\mathbb{R}}
\newcommand{\EE}{\mathbb{E}}
\newcommand{\Lcal}{\mathcal{L}}
\newcommand{\bx}{\mathbf{x}}
\newcommand{\bg}{\mathbf{g}}
\newcommand{\bu}{\mathbf{u}}
\newcommand{\bb}{\mathbf{b}}

\newcommand{\btau}{\boldsymbol{\tau}}
\newcommand{\norm}[1]{\left\lVert#1\right\rVert}

\newcommand{\fit}[1]{\adjustbox{max width=\linewidth}{#1}}

\usepackage[hidelinks]{hyperref}

\journal{Image and Vision Computing}

\begin{document}
\shortcites{shan_diffusion-based_2023,wang_lvmcn_2025,tang_gen-obt_2022}
\def\floatpagepagefraction{1}
\def\textpagefraction{.001}

\begin{frontmatter}

\title{A Diffusion-Based Kinematics and Dynamics Model for 3D Sign Language Production}

\author[1]{Emanuele Colonna}
\author[2]{Moises Diaz}
\author[1]{Gennaro Vessio\corref{cor1}}
\ead{gennaro.vessio@uniba.it}
\author[2]{Miguel Angel Ferrer}
\author[1]{Giovanna Castellano}

\cortext[cor1]{Corresponding author}

\affiliation[1]{organization={Department of Computer Science, University of Bari Aldo Moro}, city={Bari}, country={Italy}}

\affiliation[2]{organization={Institute for Technological Development and Innovation in Communications, University of Las Palmas de Gran Canaria}, city={Las Palmas de Gran Canaria}, country={Spain}}

\begin{abstract}
Sign Language Production (SLP) generates continuous signing from a linguistic input and drives accessibility technologies such as signing avatars. Current generators, diffusion models included, regress joint coordinates as an unstructured vector, so nothing ensures that the generated body respects anatomy or the mechanics of movement. Physics has entered SLP only through geometric loss terms, whose benefit for intelligibility has not been verified, and the dynamics of the signing body has not been modelled. We present PIDiffSign, a diffusion model informed by physics at two levels. For kinematics, a forward-kinematics head outputs bone directions and lengths and composes them into an anatomically exact skeleton. For dynamics, a rigid-body model of the upper body with anthropometric parameters supports two training priors. A physical-plausibility suite that includes two laws of motor control evaluates the generated signing. Under a controlled protocol, the head lowers back-translation word error rate by about five points on PHOENIX14T and CSL-Daily, whereas geometric penalties have no measurable effect; a control experiment traces the gain to the kinematic-chain representation and not to bone-length conservation. The dynamics priors gain about one further point over matched non-physical controls on PHOENIX14T, but not on CSL-Daily. Generated signing follows the laws of motor control as closely as real signing but carries markedly less kinetic energy. The exact skeletons can drive rigged avatars directly, and the suite offers a physical benchmark for future SLP systems.

\end{abstract}

\begin{keyword}
Sign Language Production \sep Diffusion Models \sep Physics-Informed Learning \sep Rigid-Body Dynamics
\end{keyword}

\end{frontmatter}

\section{Introduction}
\label{sec:intro}

Sign languages are natural languages, with their own grammar and lexicon, expressed through the coordinated movement of the hands, arms, body and face. Making spoken and written content available in sign language is a long-standing goal of assistive technology for deaf communities. Sign Language Production (SLP), the automatic generation of continuous signing from a linguistic input, is the part of this effort that produces the visible output, usually a 3-D skeleton that drives a signing avatar or a video renderer. Most SLP systems follow a pipeline that first translates speech or text into glosses and then translates glosses into motion. Glosses are written labels, conventionally in capitals of the spoken language, that name each sign in the order in which it is performed; they transcribe the lexical content of a signed sentence but not its form. This work addresses the gloss-to-pose step. Its output must meet two requirements: it must be understood as the intended sentence, and it must look like the movement of a real human body.

Current methods address the first requirement more directly than the second. Progressive Transformers~\citep{saunders_progressive_2020} cast gloss-to-pose generation as the regression of joint coordinates, and later work extended this formulation with adversarial training~\citep{saunders_adversarial_2020}, mixture-density outputs~\citep{saunders_continuous_2021} and non-autoregressive decoding~\citep{zelinka_neural_2020}. Diffusion models are now the prevailing generative approach~\citep{Baltatzis_2024_CVPR,xie_g2p-ddm_2024}, with GCDM~\citep{tang_gcdm_2024}, SignDiff~\citep{fang_signdiff_2025} and Sign-IDD~\citep{xie_sign-idd_2025} among the most recent. These models keep the same output, one 3-D position per joint and frame. Because this representation carries no notion of a skeleton, bones can change length from frame to frame, joints can reach impossible configurations and motion can be jittery, failure modes that motivate recent SLP work on bone structure~\citep{xie_sign-idd_2025,wu_geometry_2025}.

Physical knowledge has so far entered SLP only to a limited extent. The methods that target anatomical plausibility add geometric terms to the training loss, such as bone-length and bone-direction losses~\citep{xie_sign-idd_2025,wu_geometry_2025}, in the spirit of the soft constraints of physics-informed neural networks (PINNs)~\citep{raissi_physics-informed_2019,karniadakis_physics-informed_2021}. We follow the same strategy in our baselines, with penalties on bone length, joint angles and acceleration (Section~\ref{sec:exp:kin}). Three problems remain open. First, it is not known whether such penalties make the generated signing more intelligible, because SLP results are reported under protocols that differ in back-translator, pose representation and model-selection split, without an estimate of the variation between training runs. Second, no SLP method models the dynamics of the signing body, that is, the masses, inertia and forces that shape real movement, although these are standard tools of biomechanics~\citep{winter_biomechanics_2009}. Third, there is no established measure of the physical plausibility of generated signing: evaluation relies on back-translation scores and pose distances, which say nothing about the energy of a movement or about its agreement with the known laws of human motor control.

In this paper, we introduce PIDiffSign, a diffusion model for gloss-to-pose SLP that ranks among the first methods guided by physics beyond geometric loss terms and the first to incorporate rigid-body dynamics and human motor control principles. Physical knowledge enters the model at three levels, each addressing one of the stated problems, and is evaluated under a controlled protocol on two public corpora, PHOENIX14T~\citep{forster_extensions_2014} (German Sign Language) and CSL-Daily~\citep{zhou_improving_2021} (Chinese Sign Language). Our contributions include the following:
\begin{enumerate}
\item \textbf{Kinematics in the output representation.} A forward-kinematics (FK) output head predicts the position of a root joint, one unit direction per bone and frame and one length per bone, and composes them into joint positions (Figs.~\ref{fig:method} and~\ref{fig:fk}). Bone lengths are constant by construction, as in hard-constrained PINNs~\citep{lu_physics-informed_2021} and in the kinematic layers used for motion retargeting~\citep{villegas_neural_2018,pavllo_modeling_2020}. The head replaces the final linear layer of a denoiser at the same parameter count.
\item \textbf{Dynamics in the training objective.} We model the upper body as a rigid-body system with segment masses, centres of mass and radii of gyration from anthropometric tables~\citep{de_leva_adjustments_1996}. On this model we define two priors: an \emph{inertial-metric} velocity loss, which measures the kinetic energy of the error motion, and a \emph{Newton--Euler} loss, which compares the joint moments, gravity included, required by generated and real motion. Each prior is paired with a Euclidean control of the same differential order and without masses, which separates the contribution of the physics from that of supervising motion.
\item \textbf{Physics in the evaluation.} A physical-plausibility suite measures bone-length consistency, kinetic energy, speed, joint moments and jerk after the low-pass differentiation that biomechanics applies before inverse dynamics~\citep{winter_biomechanics_2009,savitzky_smoothing_1964}. It also tests two laws of human motor control: the lognormality principle of the Kinematic Theory of rapid human movements~\citep{plamondon_kinematic_1995,ferrer_idelog_2020,plamondon_open_2023}, under which a fluent speed profile is a sum of lognormal strokes, and the two-thirds power law between speed and curvature~\citep{lacquaniti_law_1983}.
\item \textbf{A controlled evaluation protocol.} Published SLP results are obtained with different back-translators, pose representations and model-selection splits, usually from a single training run, so they cannot be compared directly (Section~\ref{sec:related:slp}). We fix these three factors, report the back-translation scores of the ground-truth poses (ceiling) and of a static mean pose (floor), and train every configuration several times, so that differences are read against the spread between independent runs. Under this protocol we retrain Sign-IDD, the strongest public baseline, and insert the FK head into it.
\end{enumerate}

The rest of the paper is organised as follows. Section~\ref{sec:related} reviews related work. Section~\ref{sec:method} describes the model, the rigid-body model of the upper body and the dynamics priors. Section~\ref{sec:exp} presents the experimental setup, the evaluation metrics, including the physical-plausibility suite, and the results. Section~\ref{sec:discussion} discusses their implications and limitations, and Section~\ref{sec:conclusion} concludes and outlines open problems.

\section{Related Work}
\label{sec:related}

The present study draws on three lines of research. We first review SLP and the way it is evaluated (Section~\ref{sec:related:slp}), then the SLP methods that already constrain the skeleton (Section~\ref{sec:related:anat}). Finally, the utilisation of physical knowledge in human motion is explored, both in terms of its generation and its assessment through the laws of motor control (Section~\ref{sec:related:phys}).

\subsection{Sign language production}
\label{sec:related:slp}

SLP methods are commonly grouped by the generative model that produces the pose sequence. Regression-based methods predict poses directly. Progressive Transformers~\citep{saunders_progressive_2020} introduced gloss-to-pose and text-to-pose generation with an autoregressive Transformer, and later work added adversarial training~\citep{saunders_adversarial_2020}, mixture-density outputs~\citep{saunders_continuous_2021} and non-autoregressive decoding~\citep{zelinka_neural_2020}. A regression loss averages over the many poses compatible with a sentence, which tends to produce over-smoothed motion. Diffusion models avoid this averaging by sampling from the conditional distribution, and they are now the dominant approach: G2P-DDM~\citep{xie_g2p-ddm_2024}, GCDM~\citep{tang_gcdm_2024}, the 3-D avatar generator of \citet{Baltatzis_2024_CVPR}, SignDiff~\citep{fang_signdiff_2025}, Sign-IDD~\citep{xie_sign-idd_2025} and Sign-MExD~\citep{shen_signmexd_2025}. More recent work moves towards multilingual and avatar-based generation, with discrete motion tokens~\citep{zuo_soke_2025,symeonidis_m3t_2026}, flow matching on sparse keyframes~\citep{low_signspark_2026}, diffusion conditioned on text or phonological features~\citep{he_textdriven_2025,feng_text2sign_2025,hong_phonology_2026} and generative models of 3-D or photorealistic signers~\citep{fang_stablesigner_2025,lakhal_diverse_2025,alghamdi_tamaththul3d_2026}. \citet{rastgoo_slp_review_2021} review earlier work.

The evaluation of SLP is less settled than its modelling. Generated poses cannot be compared with a single reference, so most works use back-translation: a frozen sign language translation model reads the generated poses, and its output is scored against the source sentence with BLEU or word error rate (WER)~\citep{saunders_progressive_2020}. This score depends on three choices: the back-translation model, the pose representation it receives and the data split used to select checkpoints. Published methods differ in these choices and usually report a single training run, so their numbers are not directly comparable, and how closely back-translation agrees with human judgement is itself debated~\citep{jiang_meaningful_2025,cory_backtranslation2_2026}. The first SLP challenge was organised in 2025 partly to provide a common evaluation~\citep{walsh_slrtp2025_2025}. Back-translation also measures only whether the content can be recovered, and not whether the skeleton is anatomically valid or the motion physically plausible. We keep back-translation as the semantic measure, but we fix the three choices above for every method and retrain Sign-IDD under the same conditions instead of comparing with its published numbers. We complement it with a physical-plausibility suite (Section~\ref{sec:exp:metrics}) that measures what back-translation does not.

\subsection{Anatomical constraints in sign language production}
\label{sec:related:anat}

Few SLP works constrain the skeleton. Sign-IDD~\citep{xie_sign-idd_2025} gives the denoiser iconicity-disentangled bone features, namely direction and length, and supervises bone attributes during training, but its output is still a set of joint coordinates. \citet{wu_geometry_2025} add bone-length, bone-pose and parent-relative losses to a text-to-pose generator and bring its bone statistics closer to the ground truth. The 3-D avatar generator of \citet{Baltatzis_2024_CVPR} regresses SMPL-X joint rotations, which keep bone lengths fixed along the kinematic tree, but it requires the body-model parameters of every training sequence, obtained through a fitting pipeline. Except for the rotation-based avatar, all these methods constrain the skeleton through a penalty on joint coordinates: the model is encouraged to satisfy the constraint but can still violate it. None of them models the dynamics of the signer.

\begin{table}[H]
\caption{How related SLP methods treat the skeleton and the motion. \emph{Output}: quantity predicted by the generator. \emph{Bone lengths}: how their constancy is obtained. \emph{Dynamics}: whether the inertia of the body segments enters training.}
\label{tab:related}
\centering
\small
\setlength{\tabcolsep}{5pt}
\fit{\begin{tabular}{lccc}
\toprule
Method & Output & Bone lengths & Dynamics \\
\midrule
Progressive Transformers~\citep{saunders_progressive_2020} & joint coordinates & none & -- \\
Sign-IDD~\citep{xie_sign-idd_2025} & joint coordinates & supervised & -- \\
\citet{wu_geometry_2025} & joint coordinates & loss & -- \\
\citet{Baltatzis_2024_CVPR} & SMPL-X rotations & exact (body model) & -- \\
PIDiffSign & bone directions and lengths & exact (FK head) & priors + metrics \\
\bottomrule
\end{tabular}}
\end{table}

Table~\ref{tab:related} summarises the differences. Our PIDiffSign works directly on the keypoint annotations that SLP corpora provide and predicts bone directions and lengths, from which a forward-kinematics layer computes the coordinates. Bone lengths are then constant by construction, without a body model and without a penalty. In addition, we describe the upper body as a set of rigid segments and derive dynamics priors from it, which no previous SLP method does.

\subsection{Physical knowledge in human motion}
\label{sec:related:phys}

Outside SLP, physical knowledge enters motion generation in two ways~\citep{khani_motion_survey_2025}. The first is through a simulator. PhysDiff~\citep{yuan_physdiff_2023} projects intermediate diffusion samples through a physics simulator to remove foot sliding and ground penetration, and physics-based motion capture methods such as PhysCap~\citep{shimada_physcap_2020} and the contact-aware estimator of \citet{rempe_contact_2020} fit poses to rigid-body dynamics. These methods address whole-body locomotion, where contact and gravity dominate. Signing is performed by the upper body without ground contact, and what matters is the inertia of the arm chain and the articulation of the fingers. The second way is through the loss. Motion diffusion models~\citep{tevet_human_2023,zhang_motiondiffuse_2022} commonly add velocity or foot-contact terms on joint coordinates.

The choice between a loss and the representation is a question of soft versus hard constraints. Physics-informed neural networks add the residual of a governing equation to the loss~\citep{raissi_physics-informed_2019,karniadakis_physics-informed_2021}, whereas hard-constrained formulations build the constraint into the output and are reported to converge better and to satisfy it exactly~\citep{lu_physics-informed_2021}. For articulated bodies, forward-kinematics layers play the second role~\citep{villegas_neural_2018,pavllo_modeling_2020}. Parametric body models are a related solution, but they require estimating body-model parameters through fitting pipelines rather than learning from raw keypoints. Our dynamics priors are soft constraints in the sense above, but they weight the motion by segment inertia and compare generalised forces, instead of acting on joint coordinates. The FK head is the hard counterpart for the kinematics. Our results agree with the preference for hard constraints and add that the gain also appears when the constraint is removed and only the coordinates are kept (Section~\ref{sec:exp:kin}). Because we operate on the standard 3-D annotations of SLP corpora, we leave an end-to-end comparison with parametric rotation heads to future work.

Physical knowledge can also serve to assess motion rather than to generate it. Motor control provides regularities that a plausible movement obeys. The Kinematic Theory of rapid human movements models the speed profile of a movement as the output of a neuromuscular system with a lognormal impulse response, and the Sigma-Lognormal model decomposes complex movements into overlapping lognormal strokes, the elementary movements between two minima of the speed profile~\citep{plamondon_kinematic_1995}. Extraction algorithms such as iDeLog~\citep{ferrer_idelog_2020} and the synthesis of 3-D air-writing and gestures~\citep{ferrer_synthesis_2023} have made the theory a practical tool for assessing movement fluency~\citep{diaz_improving_2020}, with applications reviewed by \citet{plamondon_lognormality_book_2021} and \citet{plamondon_open_2023}. The two-thirds power law~\citep{lacquaniti_law_1983} relates tangential speed to curvature in drawing and reaching. We use both laws as evaluation criteria for generated signing, by comparing how closely generated and real sequences follow them (Section~\ref{sec:method:suite}). Neither has been used to evaluate generated signing before.

\section{Method}
\label{sec:method}

\begin{figure}[H]
\centering
\includegraphics[width=\textwidth]{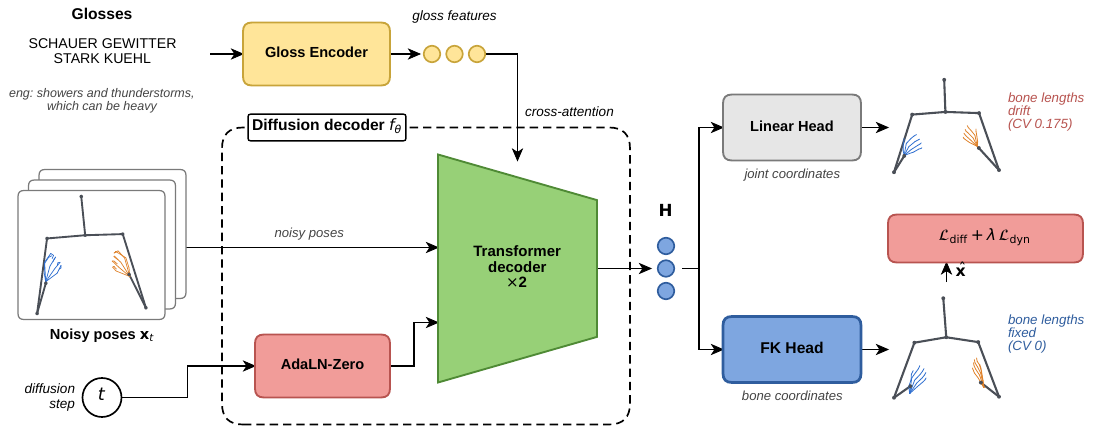}
\caption{Overview of PIDiffSign. A diffusion decoder turns noisy poses, the diffusion step and the glosses into per-frame states $\mathbf{H}$. A linear head maps them to joint coordinates, so bone lengths drift from frame to frame (faint skeletons: several frames of the same arm). The forward-kinematics head maps them to bone coordinates and keeps the lengths fixed. Its output $\hat\bx$ feeds the denoising loss and the dynamics priors.}
\label{fig:method}
\end{figure}

PIDiffSign introduces physical knowledge at three levels (Fig.~\ref{fig:method}). Kinematics enters the output representation through the FK head (Section~\ref{sec:method:fk}). Dynamics enters the training objective through a rigid-body model of the upper body (Section~\ref{sec:method:dyn}) and the priors built on it (Section~\ref{sec:method:dynloss}). The same rigid-body model underlies the physical-plausibility metrics used for evaluation (Section~\ref{sec:exp:metrics}). All components are added to a standard conditional diffusion backbone (Section~\ref{sec:method:backbone}).

\subsection{Problem and diffusion backbone}
\label{sec:method:backbone}

Let $\bg=(g_1,\dots,g_S)$ be a gloss sequence and $\bx\in\RR^{T\times J\times 3}$ the corresponding upper-body pose sequence of $T$ frames, with $J{=}50$ joints: neck (the root), head, shoulders, elbows and wrists, and $21$ joints per hand. The skeleton is a tree with $K{=}J{-}1$ bones $(p_k,c_k)$ oriented away from the root; a binary mask marks valid frames. We learn $p_\theta(\bx\mid\bg)$ with a conditional diffusion model~\citep{ho_denoising_2020} that predicts the clean sample: the network $f_\theta(\bx_t,t,\bg)$ receives a noisy sequence $\bx_t=\sqrt{\bar\alpha_t}\bx+\sqrt{1-\bar\alpha_t}\,\boldsymbol\epsilon$ and returns $\hat\bx=f_\theta(\bx_t,t,\bg)$, trained with the standard denoising loss, the frame-masked squared error $\Lcal_{\mathrm{diff}}=\EE\,\norm{\hat\bx-\bx}^2$. We also test an $\ell_1$ version of this loss, $\EE\,\norm{\hat\bx-\bx}_1$, which Sign-IDD uses (Section~\ref{sec:exp:ops}). The encoder is a two-layer Transformer over gloss embeddings. The denoiser is a two-layer Transformer decoder that self-attends over frames, cross-attends to the gloss encoding and is conditioned on $t$ through AdaLN-Zero~\citep{peebles_scalable_2023}. We use a squared-cosine noise schedule with $1000$ training steps and DDIM sampling~\citep{song_denoising_2021} with $50$ steps. All physical quantities below are computed in the stored (metric-like) units of the dataset.

\subsection{Forward-kinematics output head}
\label{sec:method:fk}

\begin{figure}[H]
\centering
\includegraphics[width=\textwidth]{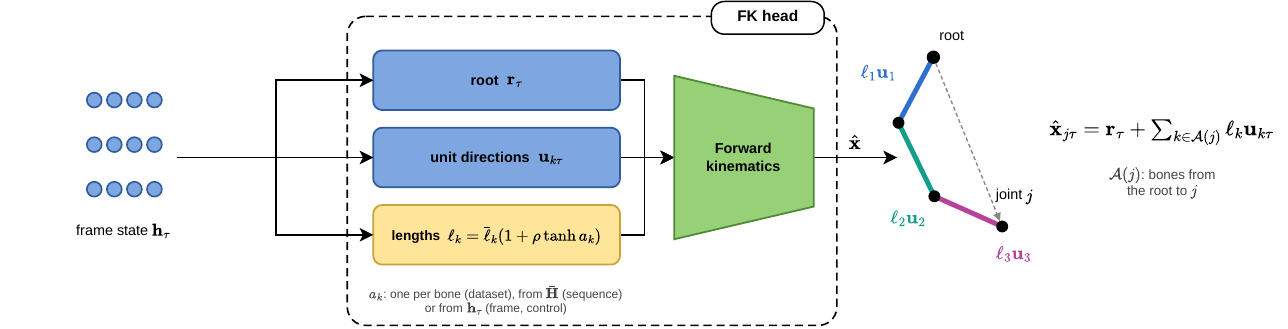}
\caption{The FK head. From the frame state it predicts a root, a unit direction per bone and frame and a length per bone; a joint is the root plus the sum of the bone vectors on its path. The three length variants (dataset-level, per-sequence and per-frame) differ only in where the length is computed.}
\label{fig:fk}
\end{figure}

In a standard denoiser, a linear output head maps the decoder states $\mathbf{H}\in\RR^{T\times d}$ to $J\times3$ joint coordinates, so any kinematic property of the output has to be learned or encouraged by a loss. The FK head (Fig.~\ref{fig:fk}) replaces this map with the description of the pose that an articulated body naturally admits: where the root is, in which direction each bone points and how long each bone is. For every frame $\tau$, the head predicts from the normalised state $\mathbf{h}_\tau$ the root position $\mathbf{r}_\tau\in\RR^3$ and, for every bone $k$, a unit direction $\bu_{k\tau}$, obtained by normalising a learned residual added to the mean direction of that bone in the training set. Each bone also receives a length $\ell_k=\bar\ell_k\,(1+\rho\tanh a_k)$, which can deviate from the reference length $\bar\ell_k$ measured on the training set by at most a fraction $\rho=0.5$ of it. Joint positions then follow by forward kinematics,
\begin{equation}
  \hat\bx_{j\tau}=\mathbf{r}_\tau+\sum_{k\in\mathcal{A}(j)}\ell_k\,\bu_{k\tau},
  \label{eq:fk}
\end{equation}
where $\mathcal{A}(j)$ is the set of bones on the path from the root to joint $j$. Since the tree is fixed, Eq.~\eqref{eq:fk} is a single product with a constant ancestor matrix, and the head costs about as much as the linear layer it replaces.

The length coefficient $a_k$ controls how bone lengths may vary, and we consider three variants of the head that differ only in this respect. With \emph{dataset-level} lengths, $a_k$ is a single learned parameter per bone, shared by all sequences of the training corpus (PHOENIX14T or CSL-Daily), so every sequence is performed by the same skeleton. With \emph{per-sequence} lengths, $a_k$ is predicted from the temporal average of the decoder states, so each sequence has its own skeleton, constant over time. With \emph{per-frame} lengths, $a_k$ is predicted from the state of each frame, so the lengths may change from frame to frame. The first two variants conserve bone lengths by construction. The third uses the same chain coordinates and the same number of parameters as the per-sequence variant but gives no conservation guarantee, and it therefore serves as a control that separates the effect of the coordinates from that of the constraint: if conserving bone lengths is what improves intelligibility, the control should be worse than the other two variants; if it is not, the gain comes from the chain coordinates.

\subsection{Rigid-body model of the upper body}
\label{sec:method:dyn}
A sequence of joint positions describes where the body is, but not what it costs to move it. Real signing is produced by muscles that accelerate segments of very different mass: moving the whole arm requires far larger forces than curling a finger, and a generator should be judged accordingly. To introduce this knowledge, we describe the upper body as a chain of rigid segments, one per bone, following the standard practice of biomechanics~\citep{winter_biomechanics_2009}. Each segment $k$ has a mass $m_k$, expressed as a fraction of body mass, a centre of mass located at a fraction $c_k$ of its length from the proximal joint, and a radius of gyration $\kappa_k\ell_k$ that describes how its mass is distributed around the centre. These parameters come from the anthropometric tables of \citet{de_leva_adjustments_1996}, which assign, for instance, a mass fraction of $0.027$ to the upper arm, $0.016$ to the forearm and $0.006$ to the hand; the hand mass is divided among palm and phalanges in proportion to their lengths. Since the skeleton has no trunk joints, a small mass is assigned to the clavicles to represent the shoulder girdle.

Two mechanical quantities follow from this model. Let $\bb_k=\bx_{c_k}-\bx_{p_k}$ be the vector of bone $k$ and $\mathbf{s}_k=\bx_{p_k}+c_k\bb_k$ its centre of mass. The kinetic energy of frame $\tau$ is the sum, over segments, of the translational energy of the centre of mass and the rotational energy of the segment about it,
\begin{equation}
  E_\tau=\tfrac12\sum_k m_k\Bigl(\norm{\dot{\mathbf{s}}_k}^2+\kappa_k^2\norm{\dot\bb_k}^2\Bigr),
  \label{eq:ke}
\end{equation}
which is exact for segments of constant length. The FK head guarantees this condition in its dataset-level and per-sequence variants, because there the lengths $\ell_k$ in Eq.~\eqref{eq:fk} do not depend on the frame $\tau$. The second quantity is the net moment $\btau_j$ that joint $j$ must produce to move all the segments distal to it, that is, the subtree $\mathcal{D}(j)$ hanging from $j$. Applying the Newton--Euler equations to this subtree gives
\begin{equation}
  \btau_j=\sum_{k\in\mathcal{D}(j)}\Bigl[(\mathbf{s}_k-\bx_j)\times m_k(\ddot{\mathbf{s}}_k-\mathbf{g})+m_k\kappa_k^2\,\bb_k\times\ddot\bb_k\Bigr],
  \label{eq:moment}
\end{equation}
where the first term accounts for the linear acceleration of each segment and for gravity $\mathbf{g}$, taken along the image vertical, and the second for its rotation. Both quantities are differentiable functions of the pose sequence and are computed for all segments and joints in parallel.

Velocities and accelerations must be estimated from sampled positions, and this estimate is sensitive to noise. The ground-truth poses of PHOENIX14T are lifted from 2-D keypoints, and their raw finite differences carry the jitter of the lifting: without filtering, the kinetic energy of real signing is $1.5$ times that of its low-pass filtered version and the energy of its joint accelerations $15$ times, so that the joint moments would mostly reflect noise. As is standard before inverse dynamics~\citep{winter_biomechanics_2009}, we therefore differentiate with a Savitzky--Golay filter~\citep{savitzky_smoothing_1964} (cubic polynomial, 9-frame window, cut-off of about $6$\,Hz at $25$ frames per second). The filter is applied in the same way to generated and real poses and, being a fixed linear operator, it can be used inside the training loss.

\subsection{Dynamics priors}
\label{sec:method:dynloss}
The standard denoising loss $\Lcal_{\mathrm{diff}}$ treats all joints alike: an error of one centimetre costs the same at the shoulder and at a fingertip, and the loss does not see how the error evolves over time. The dynamics priors replace this uniform measure with one derived from the rigid-body model, so that errors in the motion of the generated body are weighted by the mass and inertia of the segments involved. We define two priors, one on velocities and one on forces, both written as relative errors with respect to the real motion so that their scale does not depend on the units of the data:
\begin{align}
  \Lcal_{\mathrm{inert}}&=\frac{\sum_\tau\sum_k m_k\bigl(\norm{\dot{\hat{\mathbf{s}}}_k-\dot{\mathbf{s}}_k}^2+\kappa_k^2\norm{\dot{\hat\bb}_k-\dot\bb_k}^2\bigr)}{\sum_\tau 2E_\tau},
  \label{eq:linert}\\
  \Lcal_{\mathrm{NE}}&=\frac{\sum_\tau\sum_j\norm{\hat\btau_j-\btau_j}^2}{\sum_\tau\sum_j\norm{\btau_j}^2}.
  \label{eq:lne}
\end{align}
The inertial prior $\Lcal_{\mathrm{inert}}$ is the kinetic energy of the error motion, that is, of the difference between generated and real motion, normalised by the kinetic energy of the real motion. It is a velocity loss in which each segment counts in proportion to its mass and inertia, so a fast error of the forearm is penalised far more than an equally fast error of a phalanx. In geometric terms, it measures velocity errors in the metric that the mass matrix of the body induces on the space of configurations. The Newton--Euler prior $\Lcal_{\mathrm{NE}}$ goes one differential order further and compares the joint moments of Eq.~\eqref{eq:moment} that the generated and the real motion require, including the static load of gravity.

\begin{figure}[H]
\centering
\includegraphics[width=\textwidth]{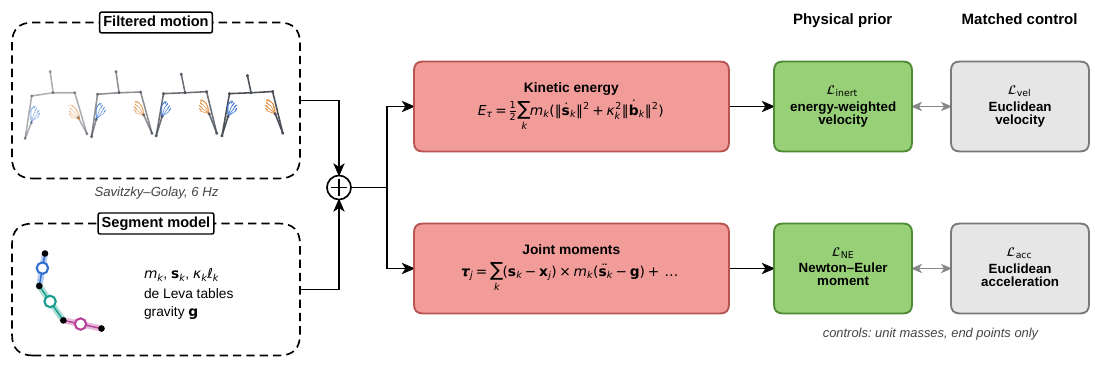}
\caption{Dynamics priors. The filtered motion of the arm, modelled as rigid segments with anthropometric masses and inertia, gives kinetic energy and joint moments. Each physical prior has a Euclidean control of the same differential order, with unit masses and end points only.}
\label{fig:dyn}
\end{figure}

A physical prior supervises the derivatives of the motion and adds physical weighting at the same time, and an improvement could come from either. To separate the two, each prior is paired with a non-physical control of the same differential order, obtained by setting every mass to $1$ and reducing every segment to its end point. The control of $\Lcal_{\mathrm{inert}}$ is the relative error of joint velocities and the control of $\Lcal_{\mathrm{NE}}$ the relative error of joint accelerations,
\begin{equation}
  \Lcal_{\mathrm{vel}}=\frac{\sum_\tau\sum_j\norm{\dot{\hat\bx}_{j\tau}-\dot\bx_{j\tau}}^2}{\sum_\tau\sum_j\norm{\dot\bx_{j\tau}}^2},\qquad
  \Lcal_{\mathrm{acc}}=\frac{\sum_\tau\sum_j\norm{\ddot{\hat\bx}_{j\tau}-\ddot\bx_{j\tau}}^2}{\sum_\tau\sum_j\norm{\ddot\bx_{j\tau}}^2},
  \label{eq:controls}
\end{equation}
both computed with the same filter (Fig.~\ref{fig:dyn}). Comparing a control with the model trained without priors tells whether supervising the motion helps at all, and comparing a physical prior with its control tells whether the physics helps beyond that. Each term is added to $\Lcal_{\mathrm{diff}}$ with weight $\lambda{=}2\times10^{-3}$, chosen so that on a trained model it is of the same order as $\Lcal_{\mathrm{diff}}$ across diffusion times.

\section{Experiments}
\label{sec:exp}

\subsection{Datasets and setup}
\label{sec:protocol}
We evaluated PIDiffSign on PHOENIX14T~\citep{forster_extensions_2014}, German Sign Language weather forecasts with 7\,096, 519 and 642 sentences for training, development and test, and on CSL-Daily~\citep{zhou_improving_2021}, Chinese Sign Language sentences on daily-life topics with 18\,399, 1\,077 and 1\,176 sentences.

Intelligibility was measured by back-translation with SignJoey, the Sign Language Transformer of \citet{cihan_camgoz_sign_2020}, which translates a pose sequence into glosses and spoken-language text. For PHOENIX14T we used the publicly released model. For CSL-Daily, for which no pose-based model is publicly available, we trained one with the same architecture on the ground-truth poses of the training split. Both models were kept frozen. On PHOENIX14T, poses were expressed in the native coordinates of the back-translator, namely the 3-D skeletons lifted from 2-D keypoints that are distributed with it and used by published baselines. This choice matters: an alternative 3-D lifting of the same videos, which we used in preliminary experiments, yields features that correlate with the native ones at only $r{=}0.07$ and raises the ceiling WER by $6.4$ points. On CSL-Daily we used 50 upper-body joints obtained by SMPL-X~\citep{SMPL-X:2019} forward kinematics on the public body fittings. Skeleton trees, reference bone lengths and mean bone directions were measured on each training split.

All PIDiffSign models shared the same training recipe: AdamW with learning rate $10^{-4}$, $1\,000$ warm-up steps and plateau decay (patience 7, factor 0.7), batch size 64, gradient clipping at 1.0, and, unless stated otherwise, DDIM sampling with 50 steps and no classifier-free guidance. The models of Table~\ref{tab:kin} were trained until the plateau schedule stopped them ($1\,413$--$1\,514$ epochs), and those with dynamics priors for 800 epochs, the range in which the best checkpoints of the FK models fall; on CSL-Daily we used the same number of optimisation steps. Sign-IDD was trained with its authors' code and settings, including checkpoint selection by dev DTW and DDIM sampling with 5 steps.

We compared the following configurations, all on the same backbone: the linear head; the linear head with a bone length and direction loss or with iconic bone inputs (soft kinematic constraints); the FK head with dataset-level, per-sequence or per-frame lengths (Section~\ref{sec:method:fk}); the FK head with dataset-level lengths and one dynamics prior or its Euclidean control (Section~\ref{sec:method:dynloss}); and the same models trained with the $\ell_1$ denoising loss. Our best configuration, reported in Table~\ref{tab:lit} and in the qualitative results, is the FK head with dataset-level lengths, the $\ell_1$ loss and the Newton--Euler prior.

Checkpoints were selected by dev WER, and the decoding settings (guidance scale and beam widths) were selected on dev by BLEU-4 and frozen before the test split was scored, once. To make the scores interpretable, the same back-translators were applied to the ground-truth test poses and to a static mean pose. Since even real signing is not translated perfectly, the ground truth gives the ceiling, the best score a generator can reach with a given back-translator (WER $71.97$ and BLEU-4 $11.52$ on PHOENIX14T, WER $47.80$ and BLEU-4 $6.35$ on CSL-Daily); the static pose, which carries no information, gives the floor (WER $\approx100$). All reported scores, ours and those of the baselines, should therefore be read between these two bounds rather than against zero.\footnote{Part of the evaluations used a second implementation of the SignJoey back-translator that loads the same weights. It reproduces the ceiling ($71.97$) and scores our FK model at $88.13$ WER and $8.36$ BLEU-4 instead of $88.25$ and $8.12$. Each table states which implementation its rows share, and rows are compared only within one implementation.}

Each configuration of the kinematic and dynamics comparisons was trained three times on both corpora with different random initialisations, and Sign-IDD five times, because its spread between runs is larger (Section~\ref{sec:exp:sota}). Results are read against this spread: the standard deviation of test WER ranges from $0.17$ to $0.81$ for the kinematic configurations of PIDiffSign, so we report means and standard deviations, use Welch tests to compare configurations, and treat differences between single runs below about one WER point as noise.

\subsection{Evaluation metrics}
\label{sec:exp:metrics}
\label{sec:method:suite}

The back-translator output was scored against the reference with four text metrics. WER is the word error rate of the predicted gloss sequence, that is, its edit distance from the reference glosses divided by their number (lower is better). BLEU-4~\citep{papineni_bleu_2002} is the geometric mean of the $n$-gram precisions, up to $n{=}4$, of the predicted spoken-language sentence; ROUGE-L~\citep{lin_rouge_2004} is the F-score of its longest common subsequence with the reference; and chrF~\citep{popovic_chrf_2015} is a character $n$-gram F-score, less sensitive to word inflection. We also report two pose distances to the ground truth: MPJPE, the mean Euclidean distance between generated and real joints over valid frames, and DTW, the same distance after the two sequences are aligned by dynamic time warping, normalised by the length of the alignment path.

Physical plausibility was measured on the test split with the filter of Section~\ref{sec:method:dyn}, and every ratio is generated over ground truth, so it equals $1$ for real signing. The bone CV is the coefficient of variation of each bone length within a sequence, averaged over bones; on the PHOENIX14T ground truth it is $0.022$, entirely due to the frame-varying scale of the lifting. The kinetic-energy ratio compares the total energy of Eq.~\eqref{eq:ke} of generated and real signing, and the Wasserstein-1 distance between the distributions of per-sequence log-energy compares how lively the signing is without requiring alignment. Speed ratios for body and hands, joint-moment ratios at shoulders, elbows and wrists (RMS of Eq.~\eqref{eq:moment}) and the jerk ratio (RMS of the filtered third derivative) complete the description of the motion. Two further metrics test laws of motor control on each wrist trajectory. For lognormality, the speed profile is split at its local minima into strokes, each stroke is fitted by one lognormal $D\,\Lambda(t;t_0,\mu,\sigma)$, and all strokes are refined jointly by bounded least squares with an analytic Jacobian, a velocity-only simplification of iDeLog~\citep{ferrer_idelog_2020}; we report the median reconstruction $\mathrm{SNR}_v=10\log_{10}\bigl(\sum v^2/\sum(v-\hat v)^2\bigr)$. For the two-thirds power law, we fit the exponent $\beta$ of $v=K\kappa^{-\beta}$ by regressing $\log v$ on $\log\kappa$ over the moving frames, and report $\beta$ and $R^2$.

\subsection{Comparison with published methods}
\label{sec:exp:lit}

\begin{table}[H]
\caption{Gloss-to-pose production on PHOENIX14T compared with published methods. All methods are evaluated by back-translation with the SLT model released by the authors of Sign-IDD~\citep{xie_sign-idd_2025}, on the same 3-D poses. Upper block: numbers as published (PT-base, PT-GN and D3DP-sign as re-implemented by \citet{xie_sign-idd_2025}). Lower block: our protocol (Section~\ref{sec:protocol}), mean over three training runs for PIDiffSign and five for Sign-IDD; Sign-IDD selects checkpoints by its own rule (dev DTW), and its dev scores are those of the selected checkpoints. Our BLEU-4 and ROUGE-L are computed on the spoken-language output of the back-translator, WER on its gloss output; the two ground-truth rows have the same WER and slightly different BLEU-4 and ROUGE, most likely because of decoding settings.}
\label{tab:lit}
\centering
\small
\setlength{\tabcolsep}{4pt}
\fit{\begin{tabular}{lcccccc}
\toprule
 & \multicolumn{3}{c}{Dev} & \multicolumn{3}{c}{Test} \\
\cmidrule(lr){2-4}\cmidrule(lr){5-7}
Method & BLEU-4$\uparrow$ & ROUGE$\uparrow$ & WER$\downarrow$ & BLEU-4$\uparrow$ & ROUGE$\uparrow$ & WER$\downarrow$ \\
\midrule
\multicolumn{7}{l}{\textit{As published}}\\
Ground truth & 12.13 & 29.60 & 74.17 & 11.93 & 28.98 & 71.94 \\
PT-base~\citep{saunders_progressive_2020} & 0.72 & 8.61 & 98.53 & 0.59 & 8.88 & 98.36 \\
PT-GN~\citep{saunders_progressive_2020} & 3.88 & 11.87 & 96.85 & 4.31 & 13.17 & 96.50 \\
D3DP-sign~\citep{shan_diffusion-based_2023} & 5.01 & 17.94 & 91.51 & 5.25 & 17.55 & 91.83 \\
GEN-OBT~\citep{tang_gen-obt_2022} & 8.68 & 25.21 & 82.36 & 8.01 & 23.49 & 81.78 \\
GCDM~\citep{tang_gcdm_2024} & 7.64 & 23.35 & 82.81 & 7.91 & 23.20 & 81.94 \\
Sign-IDD~\citep{xie_sign-idd_2025} & 8.93 & 27.60 & 77.72 & 9.08 & 26.58 & 76.66 \\
LVMCN~\citep{wang_lvmcn_2025} & 9.17 & 27.29 & 76.86 & 9.36 & 26.24 & 75.43 \\
\midrule
\multicolumn{7}{l}{\textit{Our protocol}}\\
Ground truth & 11.59 & 29.14 & 74.17 & 11.52 & 28.82 & 71.97 \\
Sign-IDD (retrained) & 10.24 & 27.92 & 82.47 & 10.02 & 26.94 & 81.87 \\
Sign-IDD + FK head & 10.16 & 27.63 & 82.56 & 10.11 & 26.93 & 81.94 \\
Linear head (our backbone) & 7.24 & 19.47 & 93.61 & 6.86 & 19.15 & 93.47 \\
PIDiffSign: FK head & 8.62 & 23.60 & 88.74 & 8.21 & 22.44 & 88.70 \\
PIDiffSign: FK head + $\Lcal_{\mathrm{NE}}$ & 8.81 & 24.14 & 87.66 & 8.40 & 23.03 & 87.77 \\
PIDiffSign: FK head + $\ell_1$ + $\Lcal_{\mathrm{NE}}$ & 9.21 & 25.06 & 86.62 & 9.35 & 24.23 & 86.49 \\
\bottomrule
\end{tabular}}
\end{table}

Table~\ref{tab:lit} places PIDiffSign among the published gloss-to-pose methods on PHOENIX14T. The comparison is possible because the most recent methods, from GEN-OBT to Sign-IDD and LVMCN, are evaluated with the same back-translation model and the same 3-D poses that we use: the ground-truth ceilings agree (WER $74.17$ on dev in both cases, $71.94$ against $71.97$ on test), while ground-truth BLEU-4 and ROUGE differ by less than one point, most likely because of decoding settings.

\begin{figure}[H]
\centering
\includegraphics[width=0.82\textwidth]{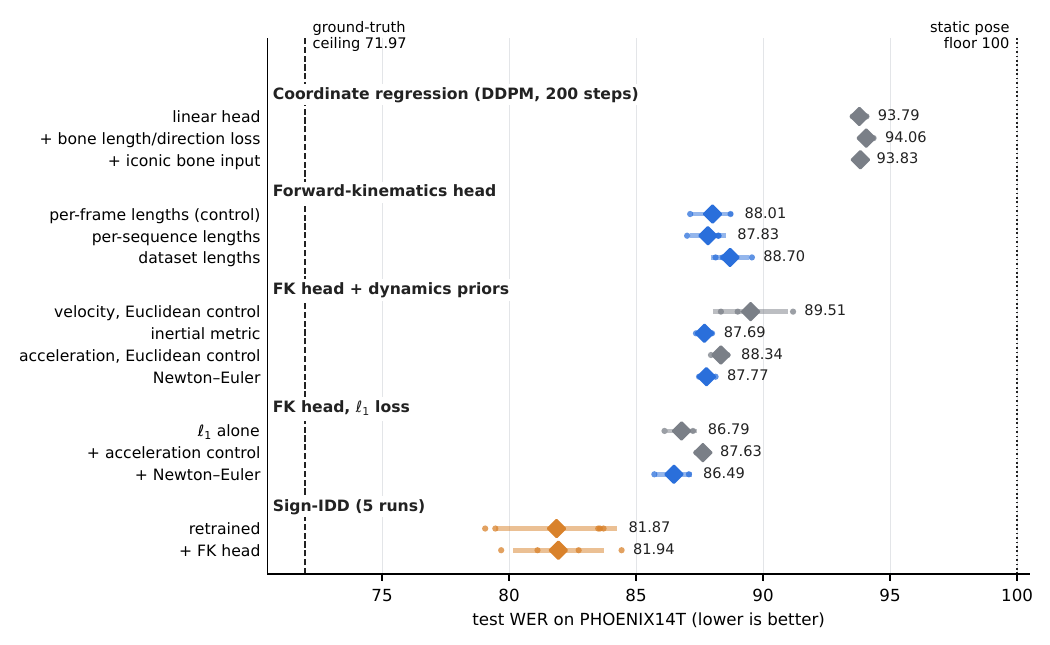}
\caption{Test WER on PHOENIX14T for every configuration. Large diamonds are means, small dots are individual training runs (three per PIDiffSign configuration, five for Sign-IDD), and bars span one standard deviation. Blue marks FK-head configurations with a physical prior or a length variant, grey marks baselines and Euclidean controls, orange marks Sign-IDD. Rows come from Tables~\ref{tab:kin}, \ref{tab:sota} and~\ref{tab:dyn}; all PIDiffSign rows use the second implementation of the back-translator (Section~\ref{sec:protocol}).}
\label{fig:overview}
\end{figure}

In BLEU-4, our best configuration (FK head, $\ell_1$ loss and Newton--Euler prior; $9.21$ on dev, $9.35$ on test) is above GEN-OBT, GCDM and the published Sign-IDD, and level with LVMCN, although our ground-truth BLEU-4 is $0.4$--$0.5$ points lower than the published one. In WER it remains about ten points behind Sign-IDD and LVMCN. This gap comes from the backbone, not from the physical components. Our backbone is deliberately small, a two-layer encoder and a two-layer decoder, so that each component can be isolated, and on it the FK head and the dynamics prior lower WER from $93.47$ to $86.49$. Inserted into Sign-IDD itself, the FK head makes every bone length exact at the same WER ($81.94$ against $81.87$; Section~\ref{sec:exp:sota}). The retrained Sign-IDD, moreover, reaches $81.87\pm2.39$ WER over five training runs with the authors' code and selection rule, five points above the published $76.66$, and its best run reaches $79.06$. A single published number can therefore lie in the tail of the distribution over training runs, and the comparisons that follow are made between configurations trained and selected under one protocol, with several runs each. The rest of this section analyses these configurations, and Fig.~\ref{fig:overview} places every PHOENIX14T configuration between the ground-truth ceiling and the static-pose floor.

\subsection{Soft kinematic penalties versus exact kinematics}
\label{sec:exp:kin}

\begin{table}[H]
\caption{Kinematics: penalties versus parametrization (test). All rows share the recipe of Section~\ref{sec:protocol}. Mean $\pm$ standard deviation over three training runs (WER, BLEU-4) or mean (other columns); PHOENIX14T rows use the second back-translator implementation; the ground-truth row uses the original one (the second gives $72.00$ WER and $11.69$ BLEU-4). The bone-loss and iconic-input rows use the sampler of their baseline (DDPM, 200 steps), the FK rows that of theirs (DDIM, 50 steps). Bone CV is the within-sequence coefficient of variation of bone length, measured on one run per configuration.}
\label{tab:kin}
\centering
\small
\setlength{\tabcolsep}{4pt}
\fit{\begin{tabular}{lcccccc}
\toprule
 & WER$\downarrow$ & BLEU-4$\uparrow$ & chrF$\uparrow$ & DTW$\downarrow$ & MPJPE$\downarrow$ & Bone CV \\
\midrule
\multicolumn{7}{l}{\textit{PHOENIX14T}}\\
Ground truth (ceiling) & 71.97 & 11.52 & 30.36 & -- & -- & 0.022 \\
Static mean pose (floor) & 100.00 & 0.00 & 0.00 & -- & -- & -- \\
Linear head, DDPM (200 steps) & 93.79 $\pm$ 0.27 & 6.85 $\pm$ 0.47 & 23.35 & 1.276 & 0.181 & 0.175 \\
\quad + bone length/direction loss & 94.06 $\pm$ 0.26 & 7.06 $\pm$ 0.23 & 23.36 & 1.282 & 0.182 & 0.202 \\
\quad + iconic bone input~\citep{xie_sign-idd_2025} & 93.83 $\pm$ 0.17 & 7.02 $\pm$ 0.41 & 23.52 & 1.279 & 0.181 & 0.161 \\
Linear head, DDIM (50 steps) & 93.47 $\pm$ 0.21 & 6.87 $\pm$ 0.28 & 23.71 & 1.236 & 0.176 & 0.175 \\
FK head, per-frame lengths (control) & 88.01 $\pm$ 0.81 & 8.42 $\pm$ 0.27 & 26.12 & \textbf{1.155} & \textbf{0.168} & 0.115 \\
FK head, per-sequence lengths & \textbf{87.83 $\pm$ 0.71} & \textbf{8.48 $\pm$ 0.27} & \textbf{26.50} & 1.168 & 0.169 & 0.000 \\
FK head, dataset-level lengths & 88.70 $\pm$ 0.76 & 8.21 $\pm$ 0.30 & 26.11 & 1.176 & 0.170 & 0.000 \\
\midrule
\multicolumn{7}{l}{\textit{CSL-Daily}}\\
Ground truth (ceiling) & 47.80 & 6.35 & -- & -- & -- & -- \\
Linear head, DDIM (50 steps) & 87.31 $\pm$ 0.47 & 3.15 $\pm$ 0.21 & 6.28 & 0.963 & \textbf{0.159} & 0.182 \\
FK head, dataset-level lengths & \textbf{82.68 $\pm$ 0.39} & \textbf{3.20 $\pm$ 0.15} & \textbf{6.43} & \textbf{0.947} & 0.160 & 0.000 \\
\bottomrule
\end{tabular}}
\end{table}

Table~\ref{tab:kin} holds the central negative and positive results. Supervising bone geometry with a length and direction loss in the style of Sign-IDD, or feeding iconicity-disentangled bone inputs to the coordinate-regression baseline, leaves WER where the baseline is, over three runs each ($94.06\pm0.26$ and $93.83\pm0.17$ against $93.79\pm0.27$; Welch $p{=}0.28$ and $0.82$). The bone loss even raises within-sequence bone variation. It regresses each bone to the length the ground truth has \emph{in that frame}, and the lifted ground truth changes scale from frame to frame.

Replacing the linear head with the FK head lowers WER on PHOENIX14T by $4.8$ to $5.6$ points at the same sampler (from $93.47\pm0.21$ to $87.83$--$88.70$; $p{=}0.004$ for the dataset-level variant), about seven times the spread between runs. The dataset-level variant does so at parameter parity, which rules out capacity as the explanation. The same change lowers WER by $4.6$ points on CSL-Daily (three runs per configuration, $87.31\pm0.47$ against $82.68\pm0.39$; $p{<}0.001$), which has a different sign language, skeleton source and back-translator. There it also lowers DTW ($0.947$ against $0.963$) at the same MPJPE.

The per-frame control tests the hypothesis stated in Section~\ref{sec:method:fk}. The decision rule was fixed before the experiment: if the control matched the per-sequence variant, the gain would be attributed to the change of coordinates and not to the conservation of bone lengths. It does match over three runs ($88.01\pm0.81$ against $87.83\pm0.71$; difference $+0.18$, Welch $p{=}0.79$), although its bone lengths vary within a sequence (bone CV $0.115$), whereas those of the per-sequence variant are constant by construction (bone CV $0$, below the $0.022$ of the lifted ground truth, whose scale changes from frame to frame). The three length variants are indistinguishable from one another: dataset-level lengths ($88.70\pm0.76$) are $0.7$--$0.9$ points behind the other two, which is within the spread ($p\geq0.22$). The hypothesis that conserving bone lengths improves intelligibility is therefore not supported. What the data support is narrower: describing the pose in the coordinates of the kinematic chain (a root, directions and lengths) gives the denoiser a much better output space than joint coordinates.

\subsection{Transfer of the FK head to Sign-IDD}
\label{sec:exp:sota}

\begin{table}[H]
\caption{FK head inside Sign-IDD~\citep{xie_sign-idd_2025} (PHOENIX14T test, mean $\pm$ standard deviation over five training runs). All rows use the authors' code, data pipeline, hyper-parameters, selection rule and sampler; only the output layer and, in the last two rows, one loss term differ. Second back-translator implementation (Section~\ref{sec:protocol}).}
\label{tab:sota}
\centering
\small
\setlength{\tabcolsep}{4pt}
\fit{\begin{tabular}{lccccccc}
\toprule
 & runs & WER$\downarrow$ & BLEU-4$\uparrow$ & chrF$\uparrow$ & DTW$\downarrow$ & KE & Bone CV \\
\midrule
Sign-IDD (retrained) & 5 & 81.87 $\pm$ 2.39 & 10.02 $\pm$ 0.61 & 28.85 & 1.129 & 0.41 & 0.172 \\
Sign-IDD + FK head & 5 & 81.94 $\pm$ 1.78 & 10.11 $\pm$ 0.56 & 28.83 & 1.117 & 0.41 & 0.000 \\
\quad + $\Lcal_{\mathrm{acc}}$ (control) & 4 & 81.17 $\pm$ 0.21 & 10.00 $\pm$ 0.22 & 28.64 & 1.106 & 0.41 & 0.000 \\
\quad + $\Lcal_{\mathrm{NE}}$ & 4 & 81.51 $\pm$ 0.86 & 10.25 $\pm$ 0.39 & 28.98 & 1.113 & 0.39 & 0.000 \\
\bottomrule
\end{tabular}}
\end{table}

To test whether the FK result is specific to our architecture, we put the head inside a different and stronger model. Sign-IDD retrained with its authors' code on our data reaches $81.87\pm2.39$ WER over five training runs, about seven points better than our FK model. The spread between runs is large, from $79.06$ to $83.72$, whereas decoding the same checkpoint again moves WER by only $0.3$--$0.6$, so the variation comes from training.

Replacing only the final linear layer with the FK head makes every bone length exact (bone CV from $0.172$ to $0$) and leaves WER, BLEU-4 and chrF unchanged within noise ($81.94\pm1.78$). Paired by random initialisation, the differences between the two models range from $-3.8$ to $+5.4$ WER points and have no consistent sign, which shows how easily a comparison based on one or two runs can suggest an effect that is not there.

This null result fits the proposed mechanism. Sign-IDD already represents the pose in chain coordinates twice. Its iconicity-disentanglement module feeds the denoiser bone lengths and directions computed from the noisy input, and its attribute-controllable decoder re-expresses the intermediate prediction as joint positions plus bone lengths and directions before the final layer. The coordinate change that the FK head brings to a plain coordinate decoder is therefore already present, and the head adds only the exactness of the constraint, which the per-frame control showed does not affect WER.

Two Sign-IDD training runs collapsed to the static mean pose and are excluded from their rows: one run of the acceleration control, whose relative acceleration error stayed at $1.0$ from the first step, and one run of the Newton--Euler condition (WER $99.2$). Since a run collapsed in each condition, including the control without any physical term, the collapse cannot be attributed to the physics term.

\subsection{Physical plausibility of generated signing}
\label{sec:exp:physics}

\begin{table}[H]
\caption{Physical-plausibility suite on PHOENIX14T test (642 sequences), computed on the scored poses of one training run per configuration; WER with the second back-translator implementation. Ratios are generated over ground truth (1 is ideal). KE: kinetic energy, Eq.~\eqref{eq:ke}. $W_1$: Wasserstein-1 distance between per-sequence log-energy distributions. Moments: RMS of Eq.~\eqref{eq:moment} at shoulder, elbow and wrist. SNR$_v$ and $\beta$: lognormal reconstruction and power-law exponent on wrist trajectories.}
\label{tab:physics}
\centering
\small
\setlength{\tabcolsep}{4pt}
\fit{\begin{tabular}{lcccccccccc}
\toprule
& WER & KE & $W_1$ & \multicolumn{2}{c}{speed} & \multicolumn{3}{c}{moment} & SNR$_v$ & $\beta$ \\
\cmidrule(lr){5-6}\cmidrule(lr){7-9}
& & ratio & log-KE & body & hands & sh. & el. & wr. & (dB) & \\
\midrule
Ground truth & 71.97 & 1.00 & 0.00 & 1.00 & 1.00 & 1.00 & 1.00 & 1.00 & 20.3 & 0.287 \\
\midrule
Linear head (DDIM 50) & 93.34 & 0.51 & 0.68 & 0.70 & 0.80 & 0.90 & 0.91 & 0.88 & 21.0 & 0.270 \\
FK, per-frame & 88.72 & 0.54 & 0.60 & 0.72 & 0.84 & 0.85 & 0.91 & 0.90 & 21.0 & 0.292 \\
FK, per-sequence & 88.25 & 0.60 & 0.50 & 0.76 & 0.86 & 0.87 & 0.94 & 0.86 & 21.1 & 0.291 \\
FK, dataset-level & 88.13 & 0.62 & 0.48 & 0.76 & 0.89 & 0.91 & 0.95 & 0.91 & 21.2 & 0.287 \\
FK + $\Lcal_{\mathrm{inert}}$ & 87.36 & 0.67 & 0.42 & 0.79 & 0.88 & 0.88 & 0.96 & 0.89 & 20.7 & 0.285 \\
FK + $\Lcal_{\mathrm{NE}}$ & 87.48 & 0.67 & 0.43 & 0.79 & 0.90 & 0.89 & 0.95 & 0.94 & 21.0 & 0.283 \\
\midrule
Sign-IDD (retrained) & 83.72 & 0.39 & 0.94 & 0.62 & 0.68 & 0.87 & 0.91 & 0.83 & 19.7 & 0.279 \\
Sign-IDD + FK head & 81.71 & 0.42 & 0.88 & 0.64 & 0.70 & 0.89 & 0.90 & 0.89 & 19.2 & 0.280 \\
\bottomrule
\end{tabular}}
\end{table}

\begin{figure}[H]
\centering
\includegraphics[width=\linewidth]{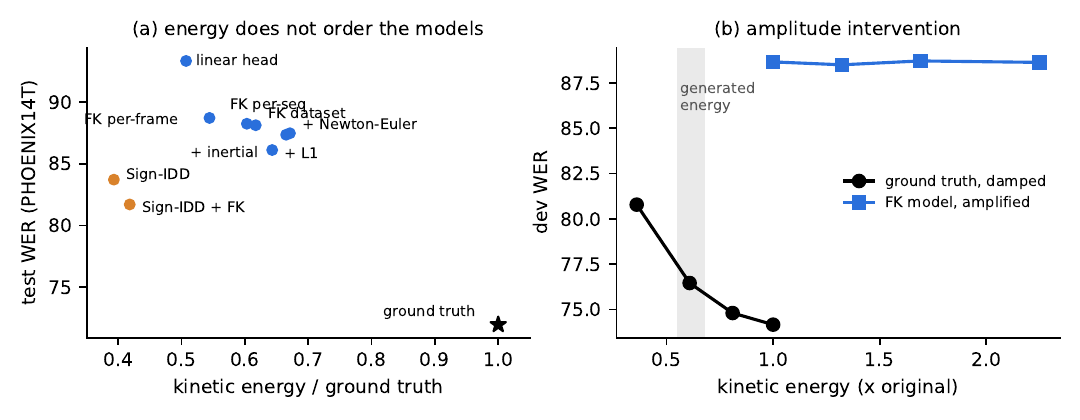}
\caption{(a) Kinetic-energy ratio and test WER of the models in Tables~\ref{tab:kin}--\ref{tab:sota} (one training run per model). Sign-IDD has the lowest energy and the lowest WER. (b) The intervention of Table~\ref{tab:energy}: damping real signing to the energy of generated signing (grey band) costs about two WER points, and amplifying generated signing changes nothing.}
\label{fig:energy}
\end{figure}

Table~\ref{tab:physics} applies the metrics of Section~\ref{sec:exp:metrics} to the scored test poses. First, generated wrist trajectories satisfy the motor-control laws about as well as real ones, for every model: speed profiles are reconstructed by sums of lognormals at a median SNR of $20$--$21$\,dB (ground truth $20.3$\,dB), and the power-law exponent is $0.27$--$0.29$ against $0.287$ for real signing, close to the classical $1/3$, so the generators do not produce jerky, non-human strokes. Second, they produce too little motion. The kinetic-energy ratio lies between $0.39$ and $0.67$ and the speed of the body joints between $0.62$ and $0.79$ of the real one, whereas joint moments, which include the static gravity load of the arm, are closer ($0.83$--$0.96$): the arms are held in plausible configurations but move through them too slowly and over too small a range. Third, the energy deficit does not rank the models by quality, since Sign-IDD has the lowest energy of all and the best WER (Fig.~\ref{fig:energy}a).

\begin{table}[H]
\caption{Amplitude intervention $\bx'=\bar\bx+k(\bx-\bar\bx)$ around the sequence mean pose (PHOENIX14T dev, FK model with dataset-level lengths, fixed decoding). Kinetic energy scales with $k^2$.}
\label{tab:energy}
\centering
\small
\setlength{\tabcolsep}{3pt}
\fit{\begin{tabular}{lccccccc}
\toprule
 & \multicolumn{4}{c}{ground truth, damped} & \multicolumn{3}{c}{generated, amplified} \\
\cmidrule(lr){2-5}\cmidrule(lr){6-8}
$k$ & 1.00 & 0.90 & 0.78 & 0.60 & 1.00 & 1.15 & 1.50 \\
KE $\times$ & 1.00 & 0.81 & 0.61 & 0.36 & 1.00 & 1.32 & 2.25 \\
WER & 74.17 & 74.81 & 76.47 & 80.79 & 88.66 & 88.50 & 88.63 \\
BLEU-4 & 11.91 & 11.62 & 11.54 & 10.15 & 8.81 & 9.03 & 8.81 \\
\bottomrule
\end{tabular}}
\end{table}

We tested the causal reading with an intervention instead of a correlation (Table~\ref{tab:energy}, Fig.~\ref{fig:energy}b). Damping real signing to the energy level of the generators ($k{=}0.78$, which scales the energy by $0.61$) costs $2.3$ WER, a sixth of the $14.5$-point gap between the FK model and the ground truth on dev. Amplifying generated motion up to $2.25\times$ its energy changes WER by less than $0.2$. Missing energy is therefore a symptom of the gap and not its main cause: the generators lack the right motion in the right place, and more of the same motion does not help. For the same reason, an energy-matching objective would be the wrong target, and we read the dynamics priors below in that light.

A second intervention asks where that place is. Replacing the generated fingers of our best configuration with the real ones, attached to the generated wrists, lowers test WER from $86.49$ to $84.27$ (three runs), and replacing the generated arms and head with the real ones, under the generated fingers, lowers it to $85.85$. The hands matter about three times more than the arms, but the two substitutions together recover less than a fifth of the $14.5$-point gap between this configuration and the ground truth on test: most of what is missing is the coordination of hand shape with arm location, not one body part.

\subsection{Dynamics priors}
\label{sec:exp:dyn}

\begin{table}[H]
\caption{Dynamics priors on the FK model with dataset-level lengths (PHOENIX14T test, mean $\pm$ standard deviation over three training runs; second back-translator implementation). Each physical prior is paired with the non-physical control of the same differential order. KE: kinetic-energy ratio of one run.}
\label{tab:dyn}
\centering
\small
\setlength{\tabcolsep}{4pt}
\fit{\begin{tabular}{lcccc}
\toprule
 & WER$\downarrow$ & BLEU-4$\uparrow$ & chrF$\uparrow$ & KE \\
\midrule
FK, dataset-level lengths (reference) & 88.70 $\pm$ 0.76 & 8.21 $\pm$ 0.30 & 26.11 & 0.62 \\
\midrule
+ $\Lcal_{\mathrm{vel}}$ (Euclidean, control) & 89.51 $\pm$ 1.49 & 7.81 $\pm$ 0.51 & 25.29 & 0.62 \\
+ $\Lcal_{\mathrm{inert}}$ (inertial metric) & \textbf{87.69 $\pm$ 0.32} & \textbf{8.44 $\pm$ 0.29} & \textbf{26.19} & 0.67 \\
\midrule
+ $\Lcal_{\mathrm{acc}}$ (Euclidean, control) & 88.34 $\pm$ 0.35 & 8.17 $\pm$ 0.11 & 26.19 & 0.65 \\
+ $\Lcal_{\mathrm{NE}}$ (Newton--Euler) & \textbf{87.77 $\pm$ 0.33} & \textbf{8.40 $\pm$ 0.17} & \textbf{26.37} & 0.67 \\
\midrule
\multicolumn{5}{l}{\textit{with an $\ell_1$ denoising loss}}\\
FK + $\ell_1$ & 86.79 $\pm$ 0.59 & 8.86 $\pm$ 0.01 & 27.08 & 0.64 \\
+ $\Lcal_{\mathrm{acc}}$ (Euclidean, control) & 87.63 $\pm$ 0.23 & 8.61 $\pm$ 0.16 & 26.67 & 0.58 \\
+ $\Lcal_{\mathrm{NE}}$ (Newton--Euler) & \textbf{86.49 $\pm$ 0.70} & \textbf{9.35 $\pm$ 0.09} & \textbf{27.58} & 0.71 \\
\bottomrule
\end{tabular}}
\end{table}

\begin{figure}[H]
\centering
\includegraphics[width=0.82\textwidth]{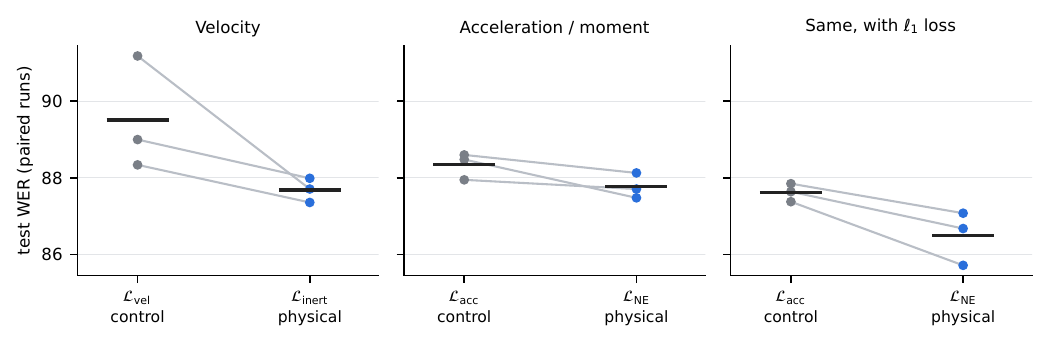}
\caption{Paired comparison of each physical prior (blue) with its Euclidean control (grey), PHOENIX14T test. Each dot is a training run; lines join runs that share the random initialisation, black bars are means. The physical prior has the lower WER in all nine pairs.}
\label{fig:pairs}
\end{figure}

Table~\ref{tab:dyn} separates the two questions of Section~\ref{sec:method:dynloss}. Supervising motion as such does not help. The Euclidean velocity and acceleration controls reach $89.51\pm1.49$ and $88.34\pm0.35$, no better than the FK reference ($88.70\pm0.76$). Supervising the same derivatives through the rigid-body model does help: the inertial-metric velocity loss reaches $87.69\pm0.32$ and the Newton--Euler moment loss $87.77\pm0.33$. Both are about one WER point below the reference, have about a third of its spread between runs, and beat their matched controls ($-1.8$ and $-0.6$; $-1.1$ in a third pair trained with the $\ell_1$ loss). The same ordering holds on dev.

The effect is modest, about a fifth of the kinematic one, and each comparison rests on three runs per configuration. The consistency is the main evidence: the physical prior has the lower WER in all nine paired runs (Fig.~\ref{fig:pairs}), although the gap of $0.6$ in the Newton--Euler pair is comparable to the control's spread. Inside Sign-IDD, the Newton--Euler prior and its control reach the same WER (Table~\ref{tab:sota}). With four runs per condition and a spread of $0.2$--$0.9$ WER, that comparison can resolve only differences of about $1.5$ points, more than the $0.6$--$1.1$ points by which the physical priors beat their controls on our backbone. Moreover, Sign-IDD already trains with an $\ell_1$ loss and a bone-attribute loss, and on our backbone the prior adds little on top of the $\ell_1$ loss ($86.49$ against $86.79$). Both physical priors also raise kinetic energy slightly ($0.67$ against $0.62$), but the intervention of Table~\ref{tab:energy} rules this out as the cause of the gain. A plausible cause is the weighting itself. On a model trained without priors, the Euclidean acceleration control places $93\%$ of its gradient on the 40 finger joints, whereas the Newton--Euler prior places less than $1\%$ there and the rest on the arm chain and the head. The arm chain carries location and movement, the parameters the gloss determines best, while the fingers of $21\%$ of the ground-truth hand frames lie at the depth of the wrist, a residue of the 2-D lifting.

On CSL-Daily the dynamics priors do not help. Over three runs each, the Newton--Euler configuration reaches $83.93\pm0.45$ and its acceleration control $83.44\pm2.01$, against $82.68\pm0.39$ for the FK model alone. The difference between prior and control ($+0.5$) is a quarter of the control's spread, and neither improves on the FK model. The two corpora do not differ in where the priors place the supervision: on CSL-Daily too, $94\%$ of the gradient of the acceleration control falls on the fingers and less than $1\%$ of that of the Newton--Euler prior. They differ in the finger targets. The CSL-Daily skeletons come from SMPL-X fits and not from 2-D lifting, and none of their hand frames is flat in depth, against $21\%$ on PHOENIX14T. Down-weighting the fingers thus removes a corrupted target on PHOENIX14T and a correct one on CSL-Daily, which is consistent with a gain on the first corpus and none on the second. We regard the dynamics result as a modest, dataset-dependent refinement and the kinematic result as the general one.

\subsection{Qualitative results}
\label{sec:exp:qual}

\begin{figure}[H]
\centering
\includegraphics[width=\textwidth]{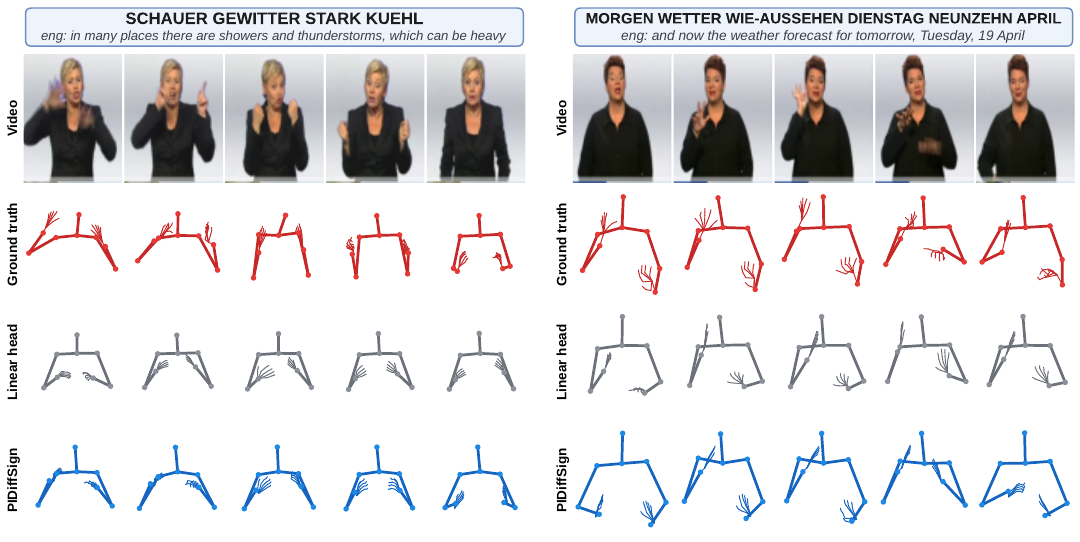}
\caption{Qualitative results on PHOENIX14T (test split) for two sentences, with their glosses. From top to bottom: video frames, ground-truth skeleton (red), linear-head backbone (grey) and PIDiffSign, FK head with $\ell_1$ loss and Newton--Euler prior (blue), at the same five uniformly spaced frames.}
\label{fig:qual_phoenix}
\end{figure}

\begin{figure}[H]
\centering
\includegraphics[width=\textwidth]{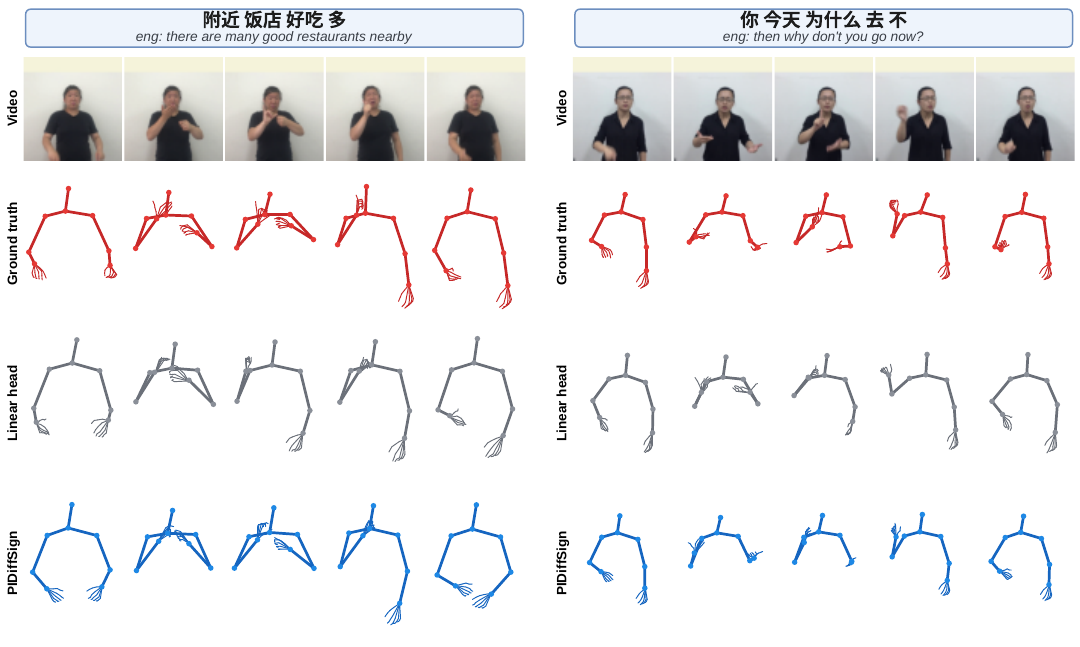}
\caption{Qualitative results on CSL-Daily (test split) for two sentences, with their glosses. From top to bottom: video frames, ground-truth skeleton (red), linear-head backbone (grey) and PIDiffSign, FK head with dataset-level lengths (blue), at the same five uniformly spaced frames.}
\label{fig:qual_csl}
\end{figure}

Figs.~\ref{fig:qual_phoenix} and~\ref{fig:qual_csl} show two test sentences of each corpus at five uniformly spaced frames, together with the ground truth and the poses generated from the glosses by the linear-head backbone and by PIDiffSign. On PHOENIX14T, PIDiffSign is the FK model with the $\ell_1$ loss and the Newton--Euler prior; on CSL-Daily, where the priors do not help, it is the FK model with dataset-level lengths. Skeletons are drawn in the image plane.

The linear head tends to keep the arms close to a rest configuration, with the hands low in front of the body, and its hands change size from frame to frame, since nothing constrains the lengths of the phalanges. PIDiffSign keeps the hand skeleton at its full size in every frame and follows the main arm movements of the reference more closely: in the second PHOENIX14T sentence it raises the arm towards the face in the same frames as the signer, and in the first CSL-Daily sentence it brings both hands to the mouth and then lowers one arm while the other stays raised, as in the reference. The figures also make the energy deficit of Section~\ref{sec:exp:physics} visible. In the first PHOENIX14T sentence the signer raises both hands to the height of the face, whereas the generated hands rise only to the chest, and in general the generated movements reach the right place with a smaller amplitude than the real ones.

\subsection{Other operating choices}
\label{sec:exp:ops}

Three choices unrelated to physics also affect WER on PHOENIX14T, each tested on the FK model with dataset-level lengths. The $\ell_1$ denoising loss of Section~\ref{sec:method:backbone} lowers test WER to $86.79\pm0.59$ over three runs (BLEU-4 $8.86\pm0.01$), consistent with the choice of Sign-IDD, whereas its ten-times larger learning rate hurts our backbone ($89.52$). Classifier-free guidance ($p_{\mathrm{drop}}{=}0.1$, $\omega{=}2$ selected on dev) lowers WER to $87.10$ and raises the kinetic-energy ratio to $0.82$. Applied to joint positions, however, guidance breaks the exactness of the kinematics (bone CV $0.017$), because the guided prediction extrapolates linearly between two sets of joint positions. Applying the same extrapolation to the parameters of the FK head instead, linearly to root and bone lengths and linearly followed by renormalisation to the bone directions, keeps every bone length exact on all 642 test sequences at the same quality: with the same checkpoint and $\omega{=}2$ again selected on dev, it gives WER $87.48$, BLEU-4 $8.63$ and DTW $1.214$, against $87.10$, $8.89$ and $1.228$ in position space (one run; the differences are within noise). Finally, fewer DDIM steps help slightly (dev WER $88.02$ at 5 steps against $88.66$ at 50).

Combined with the $\ell_1$ loss (bottom block of Table~\ref{tab:dyn}), the Newton--Euler prior again beats its acceleration control, by $1.1$ WER and $0.74$ BLEU-4 over three runs each. Against $\ell_1$ alone it is level in WER ($86.49$ against $86.79$) and better in BLEU-4 ($9.35$ against $8.86$) and chrF. This is the best configuration defined in Section~\ref{sec:protocol}.

\section{Discussion}
\label{sec:discussion}

The FK head changes two things at once. It removes the degrees of freedom that bone lengths would otherwise have, and it changes the coordinates in which the denoiser expresses a pose. The per-frame control, defined before the experiment, keeps the second change and removes the first, and over three runs it recovers the whole gain. In joint coordinates, the position of a fingertip is the sum of every bone vector above it, so an error in a proximal bone shows up as correlated errors on every distal joint, and a small change of hand shape is a small difference of large, nearly equal vectors. In chain coordinates each bone direction is a bounded unit vector that the denoiser can set independently, and the fixed matrix of Eq.~\eqref{eq:fk} does the composition. The denoiser no longer has to learn the kinematic chain and only has to learn where the bones point. This account is consistent with our results, including the unchanged WER of Sign-IDD with the FK head (Section~\ref{sec:exp:sota}): Sign-IDD already works in bone coordinates, so the head adds only the exactness of the constraint. A direct test of the account is left to future work.

Once kinematics is exact, a rigid-body description of the arm gives a better measure of motion error than the Euclidean one. The effect is about one WER point on PHOENIX14T, one fifth of the kinematic effect, and it does not act by restoring the missing kinetic energy. Its absence on CSL-Daily and inside Sign-IDD indicates that, in its current form, the dynamics prior is a refinement whose benefit depends on the data and on the model; on CSL-Daily we trace it to the quality of the finger targets (Section~\ref{sec:exp:dyn}), whereas the kinematic representation is beneficial wherever the model does not already use it.

These findings have three consequences for applications. First, signing avatars are driven by rigged skeletons with fixed bone lengths. Generators that emit joint coordinates require a retargeting step that projects each frame onto the rig and thereby alters the generated motion. The FK head removes this step, because its output is already a valid configuration of a fixed skeleton, and it does so at the cost of a linear layer and without loss of intelligibility, also when inserted into an existing model such as Sign-IDD. Second, the physical-plausibility suite gives developers of SLP systems a measure of naturalness that does not require a user study and that exposes failure modes, such as the energy deficit, which back-translation scores do not reveal. Third, the protocol shows how much SLP models vary between training runs, more than two WER points for Sign-IDD, and therefore how much evidence is needed before a single-run improvement can be trusted.

Our study also has limitations. The back-translator is a proxy for intelligibility and not a study with deaf users. The PHOENIX14T ground truth is lifted from 2-D with a frame-varying scale and intermittently flat hand depth, which limits every hand-level analysis. Every configuration of the kinematic ablation (Table~\ref{tab:kin}) and of the dynamics comparison rests on three training runs on both corpora. Sign-IDD rests on five because it varies by more than two WER points between runs, so that any comparison on it based on one or two runs can mislead. The operating choices of Section~\ref{sec:exp:ops}, guidance included, rest on a single run, except the $\ell_1$ loss. Two Sign-IDD runs collapsed to the static mean pose, one in each dynamics condition, and we report both. The segment model uses population-average anthropometry and has no trunk, and the Newton--Euler moments are those of a free upper body without contact, which suits signing but is not a validated musculoskeletal model.

\section{Conclusion}
\label{sec:conclusion}

Sign language production must generate signing that is both intelligible and physically plausible, yet current generators treat the body as an unstructured set of points, and physics has entered the task only through geometric loss terms. We have presented PIDiffSign, a diffusion model that brings physical knowledge into SLP at the level of kinematics, through a forward-kinematics output head, and of dynamics, through priors built on a rigid-body model of the upper body, and we have introduced a physical-plausibility suite that includes two laws of human motor control. Under a controlled protocol, soft kinematic penalties do not improve intelligibility, whereas the FK head makes the skeleton exact and lowers back-translation WER by about five points on PHOENIX14T and CSL-Daily. A control experiment attributes this gain to the kinematic-chain representation rather than to bone-length conservation, and inside Sign-IDD, which already uses bone coordinates, the head makes the skeleton exact without changing WER. The dynamics priors improve on matched non-physical controls by about one WER point on PHOENIX14T, but not on CSL-Daily or inside Sign-IDD. Generated signing obeys the lognormality principle and the two-thirds power law about as real signing does, but has too little kinetic energy, and an intervention shows that this deficit is not the main cause of the remaining gap. Our best configuration, the FK head with an $\ell_1$ denoising loss and the Newton--Euler prior, reaches $86.49\pm0.70$ WER and $9.35$ BLEU-4 on PHOENIX14T. Code, configurations and the physical-plausibility suite will be released.

Several problems remain open. Intelligibility is measured here by a back-translator, and a perceptual study with deaf signers is needed to confirm that exact skeletons and the dynamics priors are perceived as more natural and more understandable. The gap to the ground truth remains large, about $15$ WER points, and our analysis shows that it is not a matter of missing energy: the generators lack the right motion in the right place. Real hands on generated arms, or real arms under generated hands, recover less than a fifth of the gap, so what is missing is mainly the coordination of hand shape with arm location; of the two, the hands, whose 3-D reconstruction in PHOENIX14T is the least reliable part of the data, matter more. Corpora with accurate 3-D hand capture would allow both the dynamics priors and the motor-control analysis to be extended to the fingers. The rigid-body model can be extended with the trunk, with subject-specific anthropometry and with joint-angle limits built into the FK parametrization, moving towards a musculoskeletal description of signing. Finally, the FK head and the plausibility suite are independent of the gloss-to-pose setting and can be applied to text-to-pose generation and to the avatar-rendering stage of a complete SLP system.

\section*{CRediT authorship contribution statement}
\textbf{Emanuele Colonna:} Conceptualization, Methodology, Investigation, Validation, Writing -- original draft.
\textbf{Moises Diaz:} Conceptualization, Validation, Writing -- review \& editing.
\textbf{Gennaro Vessio:} Conceptualization, Validation, Writing -- review \& editing, Supervision.
\textbf{Miguel Angel Ferrer:} Conceptualization, Writing -- review \& editing.
\textbf{Giovanna Castellano:} Conceptualization, Writing -- review \& editing, Supervision.

\section*{Acknowledgments}
The research of Emanuele Colonna is supported by a Ph.D.~fellowship under the Italian ``D.M.\ n.\ 117/23 (2/03/2023)'', within the National Recovery and Resilience Plan, Mission 4, Component 2, Investment 3.3, Ph.D.\ project ``Study of AI techniques for efficient generation of digital humans and 3D environments'', co-supported by QuestIT S.r.l.~(CUP H91I23000690007).

We acknowledge ISCRA for awarding this project access to the LEONARDO supercomputer, owned by the EuroHPC Joint Undertaking, hosted by CINECA (Italy).

This research was partly supported by the PID2023-146620OB-I00, funded by MICIU/AEI 10.13039/501100011033 and the European Union’s ERDF program, and partly by the CajaCanaria and la Caixa (2023DIG05).

\section*{Declaration on the Use of AI-Assisted Writing Tools}
During the preparation of this work, the authors used Claude to enhance grammar and spelling accuracy. After using these tools, the authors reviewed and edited the content as needed and took full responsibility for the publication's content.

\section*{Declaration of competing interest}
The authors declare that they have no known competing financial interests or personal relationships that could have appeared to influence the work reported in this paper.

\bibliographystyle{elsarticle-harv}
\bibliography{bib}

\end{document}